\documentclass[10pt,twocolumn,letterpaper]{article}

\usepackage{cvpr}
\usepackage{times}
\usepackage{epsfig}
\usepackage{graphicx}
\usepackage{amsmath}
\usepackage{amssymb}

\usepackage[pagebackref=true,breaklinks=true,letterpaper=true,colorlinks,bookmarks=false]{hyperref}

\cvprfinalcopy 

\def\cvprPaperID{357} 

\ifcvprfinal\fi
\begin{document}

\title{Egocentric Field-of-View Localization Using First-Person Point-of-View Devices}

\author{
    Vinay Bettadapura$^{\text{1},\text{2}}$\\
    \texttt{\small vinay@gatech.edu}
  \and
	Irfan Essa$^{\text{1},\text{2}}$\\
    \texttt{\small irfan@cc.gatech.edu}
  \and
    Caroline Pantofaru$^{\text{1}}$\\
	\texttt{\small cpantofaru@google.com}
  \and
    {\small $^{\text{1}}$Google Inc., Mountain View, CA, USA}
  \\
    {\small $^{\text{2}}$Georgia Institute of Technology, Atlanta, GA, USA}
  \\
    \href{http://www.vbettadapura.com/egocentric/localization}{\small http://www.vbettadapura.com/egocentric/localization}
}

\maketitle
\thispagestyle{empty}

\begin{abstract}
We present a technique that uses images, videos and sensor data taken from first-person point-of-view devices to perform egocentric field-of-view (FOV) localization. We define egocentric FOV localization as capturing the visual information from a person's field-of-view in a given environment and transferring this information onto a reference corpus of images and videos of the same space, hence determining what a person is attending to. Our method matches images and video taken from the first-person perspective with the reference corpus and refines the results using the first-person's head orientation information obtained using the device sensors. We demonstrate single and multi-user egocentric FOV localization in different indoor and outdoor environments with applications in augmented reality, event understanding and studying social interactions.
\end{abstract}

\section{Introduction}

A key requirement in the development of interactive computer vision systems is modeling the user, and one very important question is \textit{``What is the user looking at right now?''} From augmented reality to human-robot interaction, from behavior analysis to healthcare, determining the user's egocentric field-of-view (FOV) accurately and efficiently can enable exciting new applications. Localizing a person in an environment has come a long way through the use of GPS, IMUs and other signals. But such localization is only the first step in understanding the person's FOV.

The new generation of devices are small, cheap and pervasive. Given that these devices contain cameras and sensors such as gyros, accelerometers and magnetometers, and are Internet-enabled, it is now possible to obtain large amounts of first-person point-of-view (POV) data unintrusively. Cell phones, small POV cameras such as GoPros, and wearable technology like Google Glass all have a suite of similar useful capabilities. We propose to use data from these first person POV devices to derive an understanding of the user's egocentric perspective. In this paper we show results from data obtained with Google Glass, but any other device could be used in its place.

Automatically analyzing the POV data (images, videos and sensor data) to estimate egocentric perspectives and shifts in the FOV remains challenging. Due to the unconstrained nature of the data, no general FOV localization approach is applicable for all outdoor and indoor environments. \emph{Our insight is to make such localization tractable by introducing a reference data-set}, i.e., a visual model of the environment, which is either pre-built or concurrently captured, annotated and stored permanently. All the captured POV data from one or more devices can be matched and correlated against this reference data-set allowing for transfer of information from the user's reference frame to a global reference frame of the environment. The problem is now reduced from an open-ended data-analysis problem to a more practical data-matching problem. Such reference data-sets already exist; \emph{e.g.,} Google Street View imagery exists for most outdoor locations and recently for many indoor locations. Additionally, there are already cameras installed in many venues providing pre-captured or concurrently captured visual information, with an ever increasing number of spaces being mapped and photographed. Hence there are many sources of visual models of the world which we can use in our approach.

\begin{figure*}[t]
\begin{center}
\includegraphics[width=1.7\columnwidth]{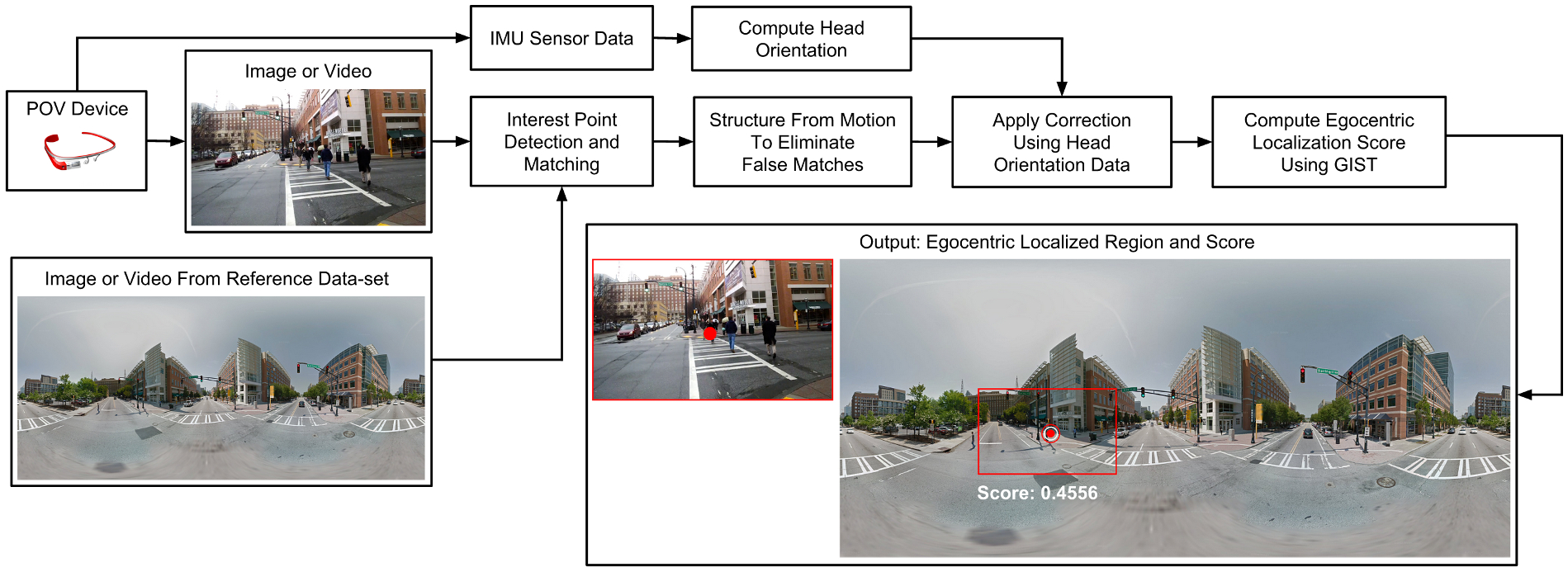}
\end{center}
\caption{An overview of our egocentric FOV localization system. Given images (or videos) and sensor data from a POV device, and a pre-existing corpus of canonical images of the given location (such as Google street view data), our system localizes the egocentric perspective of the person and determines the person's region-of-focus.}
\label{fig:block-diagram}
\end{figure*}

\noindent\textbf{Contributions:} We present a method for egocentric FOV localization that directly matches images and videos captured from a POV device with the images and videos from a reference data-set to understand the person's FOV. We also show how sensor data from the POV device's IMU can be used to make the matching more efficient and minimize false matches. We demonstrate the effectiveness of our approach across 4 different application domains: (1) egocentric FOV localization in outdoor environments: 250 POV images from different locations in 2 major metropolitan cities matched against the street view panoramas from those locations; (2) egocentric FOV localization in indoor spaces: a 30 minute POV video in an indoor presentation matched against 2 fixed videos cameras in the venue; (3) egocentric video tours at museums: 250 POV images of paintings taken within 2 museums in New York City (Metropolitan Museum of Art and Museum of Modern Art) matched against indoor street view panoramas from these museums (available publicly as part of the Google Art Project \cite{GoogleArt}); and (4) joint egocentric FOV localization from multiple POV videos: 60 minutes of POV videos captured concurrently from 4 people wearing POV devices at the Computer History Museum in California, matched against each other and against indoor street view panoramas from the museum.

\section{Related Work}

\textbf{Localization:} Accurate indoor localization has been an area of active research  \cite{SurveyLocationSystems}. Indoor localization can leverage GSM \cite{GSMLocalization}, active badges \cite{BadgeSystem}, 802.11b wireless ethernet \cite{WiFi}, bluetooth and WAP \cite{BluetoothAndWAP}, listeners and beacons \cite{Cricket}, radiofrequency \cite{RADAR} technologies and SLAM \cite{SLAM}.

Outdoor localization from images or video has also been explored, including methods to match new images to street-side images~\cite{SchindlerCity,ZamirStreetView,SchrothMobile}. Other techniques include urban navigation using a camera mobile phone \cite{UrbanNavigation}, image geo-tagging based on travel priors \cite{GeolocatingTravelPriors} and the IM2GPS system \cite{IM2GPS}.

Our approach leverages these methods for visual and sensor data matching with first-person POV systems to determine where the user is attending to.

\textbf{Egocentric Vision and Attention:} Detecting and understanding the salient regions in images and videos has been an active area of research for over three decades. Seminal efforts in the 80s and 90s focused on understanding saliency and attention from a neuroscience and cognitive psychology perspective~\cite{Treisman1980}. In the late 90s, Illti~\emph{et al.}~\cite{IttiSaliency} built a visual attention model using a bottom-up model of the human visual system. Other approaches used graph based techniques \cite{GraphSaliency}, information theoretical methods \cite{InfoMaxSaliency}, frequency domain analysis \cite{ImageSignature} and the use of higher level cues like face-detection \cite{PredictingGaze} to build attention maps and detect objects and regions-of-interests in images and video.

In the last few years, focus has shifted to applications which incorporate attention and egocentric vision. These include gaze prediction \cite{PredictGazeEgocentric}, image quality assessment \cite{ImageQuality}, action localization and recognition \cite{ActionLocalization,DailyActions}, understanding social interactions \cite{SocialInteraction} and video summarization \cite{VideoSummarization}. Our goal in this work is to leverage image and sensor matching between the reference set and POV sensors  to extract and localize the egocentric FOV.

\section{Egocentric FOV Localization}

The proposed methodology for egocentric FOV localization consists of five components: (i) POV data consisting of images, videos and head-orientation information, (ii) a pre-captured or concurrently captured reference dataset, (iii) robust matching pipeline, (iv) match correction using sensor data, and (v) global matching and score computation. An overview of our approach is shown in Figure \ref{fig:block-diagram}. Each step of the methodology is explained in detail below.

\textbf{Data collection:} POV images and videos along with the IMU sensor data are collected using one or more POV devices to construct a ``pov-dataset''. For our experiments, we used a Google Glass. It comes equipped with a 720p camera and sensors such as accelerometer, gyroscope and compass that lets us effectively capture images, videos and sensor data from a POV perspective. Other devices such as cell-phones, which come equipped with cameras and sensors, can also be used.

\textbf{Reference dataset:} A ``reference-dataset'' provides a visual model of the environment. It can either be pre-captured (and possibly annotated) or concurrently captured (i.e. captured while the person with the POV device is in the environment). Examples of such reference datasets are Google Street View images and pre-recorded videos and live video streams from cameras in indoor and outdoor venues.

\textbf{Matching:} Given the person's general location, the corresponding reference image is fetched from the reference-dataset using location information (such as GPS) and is matched against all the POV images taken by the person at that location. Since the camera is egocentric, the captured image provides an approximation of the person's FOV. The POV image and the reference image are typically taken from different viewpoints  and under different environmental conditions which include changes in scale, illuminations, camera intrinsics, occlusion and affine and perspective distortions. Given the ``in-the-wild'' nature of our applications and our data, our matching pipeline is designed to be robust to these changes.

In the first step of the matching pipeline, reliable interest points are detected both in the POV image, $I_{pov}$, and the reference image, $I_{ref}$ using maximally stable extremal regions (MSER). The MSER approach was originally proposed by \cite{MSER}, by considering the set of all possible thresholdings of an image, $I$, to a binary image, $I_B$, where $I_B(x)$=1 if $I(x)\geq{t}$ and 0 otherwise. The area of each connected component in $I_B$ is monitored as the threshold is changed. Regions whose rates of change of area with respect to the threshold are minimal are defined as maximally stable and are returned as detected regions. The set of all such connected components is the set of all extremal regions. The word extremal refers to the property that all pixels inside the MSER have either higher (bright extremal regions) or lower (dark extremal regions) intensity than all the pixels on its outer boundary. The resulting extremal regions are invariant to both affine and photometric transformations. A comparison of MSER to other interest point detectors has shown that MSER outperforms the others when there is a large change in viewpoint \cite{mikolajczyk2005comparison}. This is a highly desirable property since $I_{pov}$ and $I_{ref}$ are typically taken from very different viewpoints. Once the MSERs are detected, standard SIFT descriptors are computed and the correspondences between the interest points are found by matching them using a KD tree, which supports fast indexing and querying.

The interest point detection and matching process may give us false correspondences that are geometrically inconsistent. We use random sample consensus (RANSAC) \cite{RANSAC} to refine the matches and in turn eliminate outlier correspondences that do not fit the estimated model. In the final step, the egocentric focus-of-attention is transferred from $I_{pov}$ to $I_{ref}$. Using three of the reliable match points obtained after RANSAC, the affine transformation matrix, $\mathbf{A}$, between $I_{pov}$ and $I_{ref}$ is computed. The egocentric focus-of-attention $\mathbf{f_{pov}}$ is chosen as the center of $I_{pov}$ (the red dot in Figure \ref{fig:block-diagram}). This is a reasonable assumption in the absence of eye-tracking data. The focus-of-attention, $\mathbf{f_{ref}}$, in $I_{ref}$, is given by $\mathbf{f_{ref}}=\mathbf{Af_{pov}}$.

\textbf{Correction using sensor data:} The POV sensor data that we have allows us to add an additional layer of correction to further refine the matches. Modern cellphones and POV devices like Glass come with a host of sensors like accelerometers, gyroscopes and compasses and they internally perform sensor fusion to provide more stable information. Using sensor fusion, these devices report their absolute orientation in the world coordinate frame as a $3\times3$ rotation matrix $R$. By decomposing $R$, Euler angles $\psi$ (yaw), $\theta$ (pitch), $\phi$ (roll) can be obtained. Since Glass is capturing sensor data from a POV perspective, the Euler angles give us the head orientation information, which can be used to further refine the matches. For example, consider a scenario where the user is looking at a high-rise building that has repetitive patterns (such as rectangular windows), all the way from bottom to the top. The vision-based matching gives us a match at the bottom of the building, but the head orientation information suggests that the person is looking up. In such a scenario, a correction can be applied to the match region to make it compatible with the sensor data.

Projecting the head orientation information onto $I_{ref}$, gives us the egocentric focus-of-attention, $\mathbf{f_s}$, as predicted by the sensor data. The final egocentric FOV localization is computed as: $\mathbf{f}=\alpha\mathbf{f_s}+(1-\alpha)\mathbf{f_{ref}}$, where $\alpha$ is a value between 0 and 1 and is based on the confidence placed on the sensor data. Sensor reliability information is available in most of the modern sensor devices. If the device sensors are unreliable then $\alpha$ is set to a small value. Relying solely on either vision based matching or on sensor data is not a good idea. Vision techniques fail when the images are drastically different or have fewer features and sensors tend to be noisy and the readings drift over time. We found that first doing the vision based matching and then applying a $\alpha$-weighted correction based on the sensor data gives us the best of both worlds.

\begin{figure*}[t]
\begin{center}
\includegraphics[width=1.6\columnwidth]{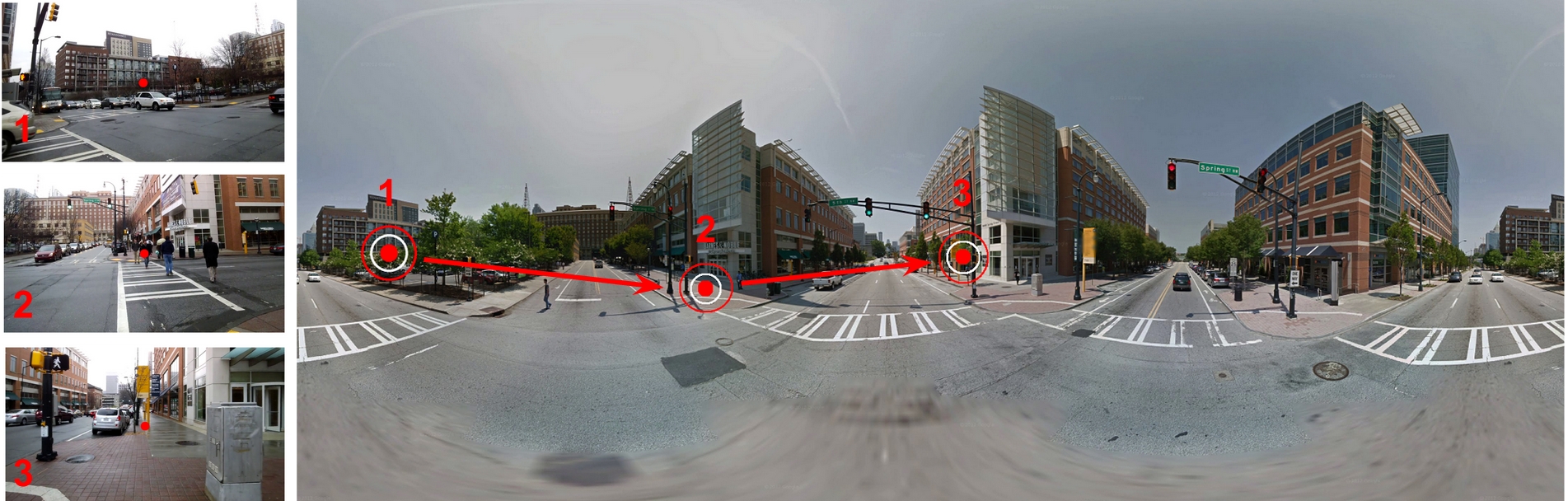}
\end{center}
\caption{Egocentric FOV localization in outdoor environments. The images on the left are the POV images taken from Glass. The red dot shows the focus-of-attention. The panorama on the right shows the localization (target symbols) and the shifts in the FOV over time (red arrows). Note the change in season and pedestrian traffic between the POV images and the reference image.}
\label{fig:outdoor_result}
\end{figure*}

\begin{figure*}[t]
\begin{center}
\includegraphics[width=1.6\columnwidth]{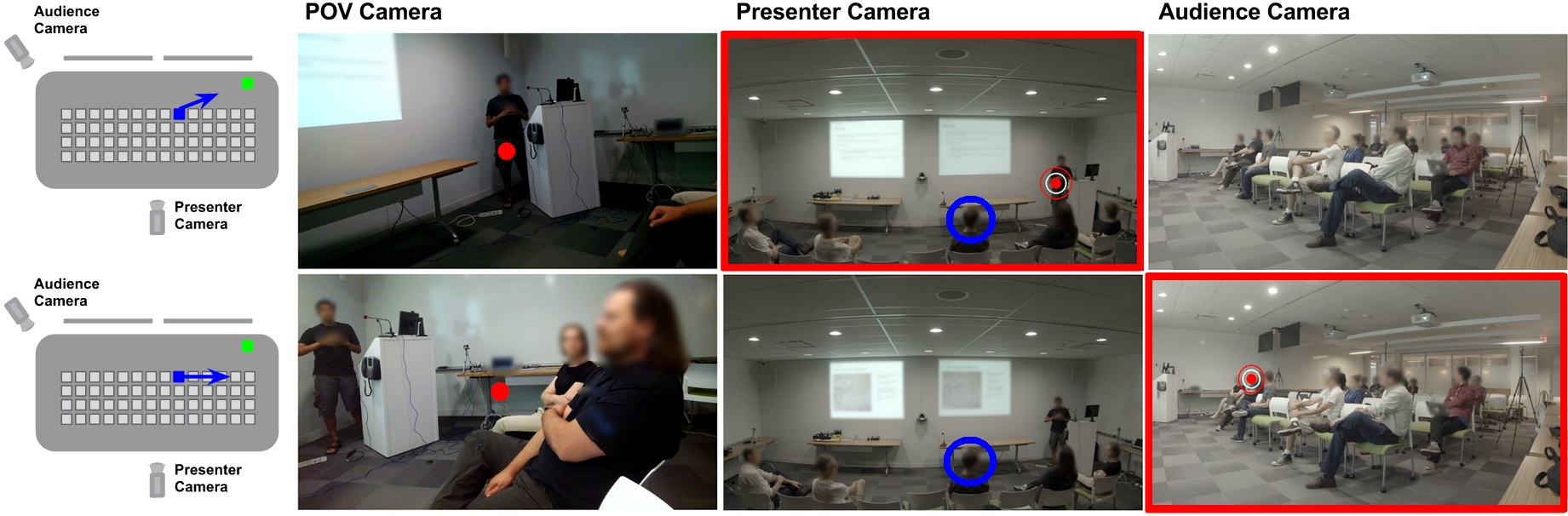}
\end{center}
\caption{Egocentric FOV localization in indoor environments. The images on the first column show the room layout. The presenter is shown in Green and the person wearing Glass is shown in Blue with his egocentric view shown by the blue arrow. The second column shows the POV video frames from Glass. The red dot shows the focus-of-attention. The third and fourth column show the presenter cam and the audience cam respectively. The localization is shown by the target symbol and the selected camera is shown by the red bounding box. The person wearing Glass is highlighted by the blue circle in the presenter camera views.}
\label{fig:indoor_result}
\end{figure*}

\begin{figure*}[t]
\begin{center}
\includegraphics[width=1.8\columnwidth]{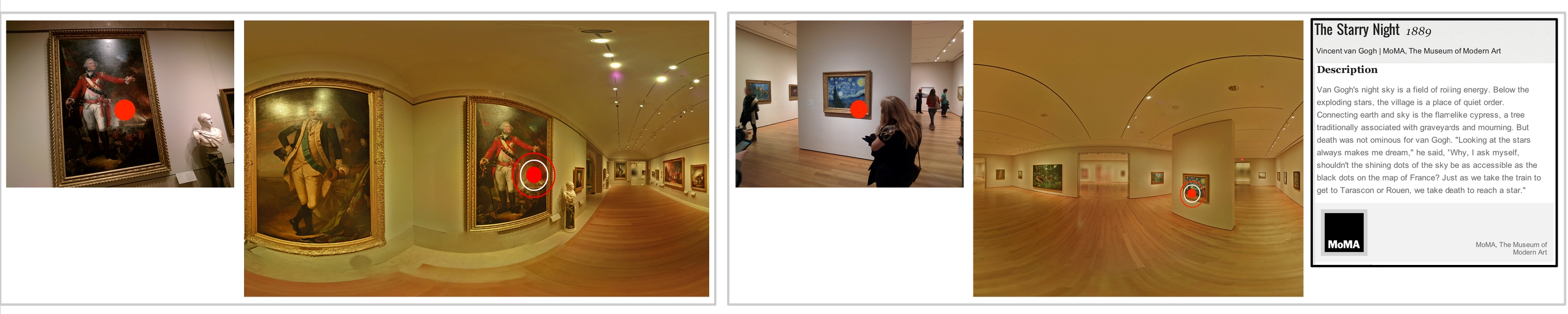}
\end{center}
\caption{Egocentric FOV localization in indoor art installations. The images on the left are the POV images taken from Glass. The red dot shows the focus-of-attention. The images to their right are panoramas from indoor streetview that correctly shows the localization result (target symbol). When available, the details of the painting are shown. This information is automatically fetched, using the egocentric FOV location as the cue. For the painting on the right (Van Gogh's ``The Starry Night''), an information card shows up and provides information about the painting.}
\vspace{-0.75em}
\label{fig:museum}
\end{figure*}

\textbf{Global Matching and Score Computation:} We now have a match window that is based on reliable MSER interest point detection followed by SIFT matching and RANSAC based outlier rejection and sensor based correction. Although this match window is reliable, it is still based only on local features without any global context of the scene. There are several scenarios in the real world (like urban environments), where we have repetitive and commonly occurring patterns and local features that may result in an inaccurate match window. In this final step, we do a global comparison and compute the egocentric localization score.

Global comparison is done by comparing the match window, $W_{ref}$ located around $\mathbf{f_s}$ in $I_{ref}$, with $I_{pov}$ (\emph{i.e.}, the red match windows of the bottom image in Figure \ref{fig:block-diagram}). This comparison is done using global GIST descriptors \cite{GIST}. A GIST descriptor gives a global description of the image based on the image's spectral signatures and tells us how visually similar the two images are. GIST descriptors $\mathbf{q_{pov}}$ and $\mathbf{q_{ref}}$ are computed for $I_{pov}$ and $W_{ref}$ respectively and final egocentric FOV localization score is computed as the $L2$-distance between the GIST descriptors: $\parallel{\mathbf{q_{pov}}-\mathbf{q_{ref}}}\parallel=\sqrt{(\mathbf{q_{pov}}-\mathbf{q_{ref}}).(\mathbf{q_{pov}}-\mathbf{q_{ref}})}$. Scoring quantifies the confidence in our matches and by thresholding on the score, we can filter out incorrect matches.

\section{Applications and Results}
To evaluate our approach and showcase different applications, we built 4 diverse datasets that include both images and videos in both indoor and outdoor environments. All the POV data was captured with a Google Glass. 

\subsection{Outdoor Urban Environments} Egocentric FOV localization in outdoor environments has applications in areas such as tourism, assistive technology and advertising. To evaluate our system, 250 POV images (of dimension 2528x1856) along with sensor data (roll, pitch and yaw of the head) was captured at different outdoor locations in two major metropolitan cities. The reference dataset consists of the 250 street view panoramas (of dimension 3584x1536) from those locations. Based on the user's GPS location, the appropriate street view panorama was fetched and used for matching. Ground truth was provided by the user who documented his point of attention in each of the 250 POV images. However we have to take into account the fact that we are only tracking the head orientation using sensors and not tracking the eye movement. Humans may or may not rotate their heads completely to look at something; instead they may rotate their head partially and just move their eyes. We found that this behavior (of keeping the head fixed while moving the eyes) causes a circle of uncertainty of radius $R$ around the true point-of-attention in the reference image. To calculate its average value, we conducted a user-study with 5 participants. The participants were instructed to keep their heads still and use only their eyes to see as far to the left and to the right as they could without the urge to turn their heads. This mean radius of their natural eye movement was measured to be 330 pixels for outdoor urban environments. Hence for our evaluation we consider the egocentric FOV localization to be successful if the estimated point-of-attention falls within a circle of radius $R=330$ pixels around the ground truth point-of-attention.

Experimental results show that without using sensor data, egocentric FOV  localization was accurate in 191/250 images for a total accuracy of 76.4\%. But when sensor data was included, the accuracy rose to 92.4\%. Figure \ref{fig:outdoor_result} shows the egocentric FOV localization results and the shifts in FOV over time. Discriminative objects such as landmarks, street signs, graffiti, logos and shop names helped in the getting good matches. Repetitive and commonly occurring patterns like windows and vegetation caused initial failures but most of them were fixed when the sensor correction was applied.

\subsection{Presentations in Indoor Spaces} There are scenarios where a pre-built reference dataset (like street view) is not available for a given location. This is especially true for indoor environments that have not been as thoroughly mapped as outdoor environments. In such scenarios, egocentric FOV localization is possible with a reference dataset that is concurrently captured along with the POV data. To demonstrate this, a 30 minute POV video along with sensor data was captured during an indoor presentation. The person wearing Glass was seated in the audience in the first row. The POV video is 720p at 30 fps. The reference dataset consists of videos from two fixed cameras at the presentation venue. One camera was capturing the presenter while the other camera was pointed at the audience. The reference videos are 1080p at 30 fps. Ground truth annotations for every second of the video were provided by the user who wore Glass and captured the POV video. So, we have 60*30 = 1800 ground truth annotations. As with the previous dataset, we empirically estimated $R$ to be 240 pixels. Experimental results show that egocentric FOV localization and camera selection was accurate in 1722/1800 cases for a total accuracy of 95.67\%. Figure \ref{fig:indoor_result} shows the FOV localization and camera selection results.

\begin{figure}
\begin{centering}
\includegraphics[width=1.0\columnwidth]{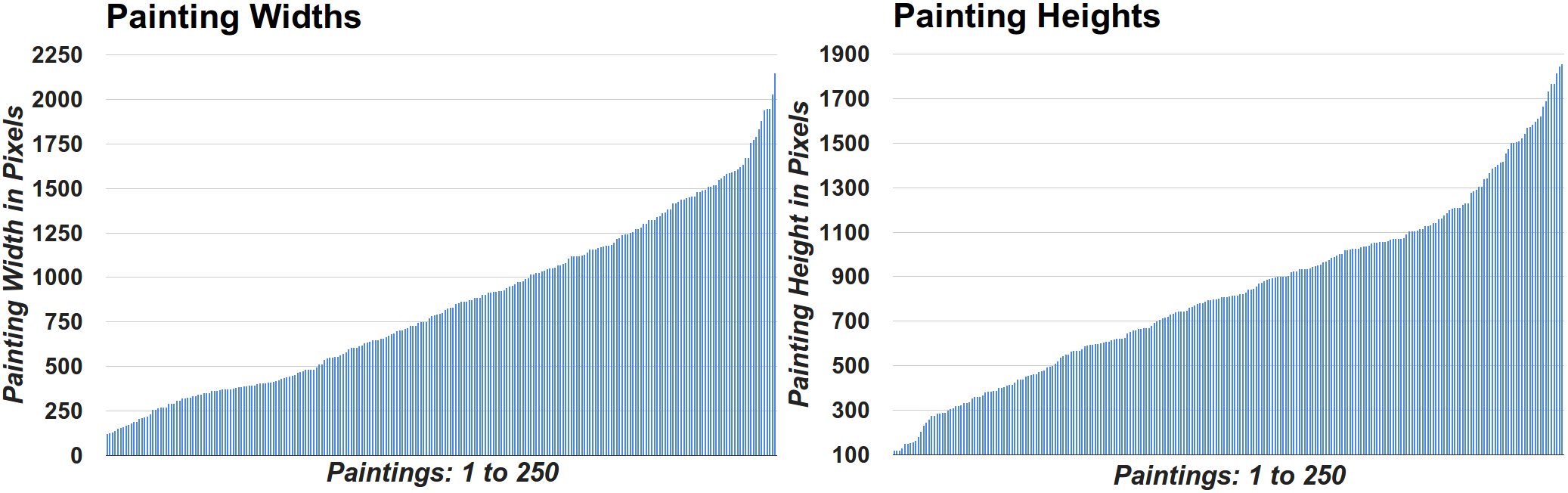} 
\par\end{centering}
\caption{\label{fig:hist}The widths and heights of the 250 paintings, sorted in ascending order based on their value. We can see that our dataset has a good representation of paintings of varying widths and heights.}
\vspace{-1.5em}
\end{figure}

\subsection{Egocentric Video Tours in Museums} Public spaces like museums are ideal environments for an egocentric FOV localization system. Museums have exhibits that people explicitly pay attention to and want to learn more about. Similar to audio-tours that are available in museums, we demonstrate a system for attention-driven egocentric video tours. Unlike in an audio tour where a person has to enter the exhibit number to hear details about it, our video tour system recognizes the exhibit when the person looks at it and brings up a cue card on the wearable device giving more information about the exhibit.

For our evaluation, we captured 250 POV images of paintings at 2 museums in New York City - The Metropolitan Museum of Art and The Museum of Modern Art. The reference dataset consists of indoor street view panoramas from these museums, made available as part of the Google Art Project \cite{GoogleArt}. Since this dataset consists of paintings, which have a fixed structure (a frame enclosing the artwork), we have a clear definition of correctness: egocentric FOV localization is deemed to be correct if the estimated focus-of-attention is within the frame of the painting in the reference image. Experimental results show that the localization was accurate in 227/250 images for a total accuracy of 90.8\%. Figure \ref{fig:hist} shows the distribution of the widths and heights of the paintings in out dataset. We can see that paintings of all widths and heights are well represented.

The Google Art panoramas are annotated with information about the individual paintings. On successful FOV localization, we fetch the information on the painting that the person is viewing and display it on Glass or as an overlay. Figure \ref{fig:museum} shows the FOV localization results and the painting information that was automatically fetched and shown on Glass.

\subsection{Joint Egocentric FOV Localization}

When we have a group of people wearing POV devices within the same event space, egocentric FOV localization becomes much more interesting. We can study joint FOV localization (i.e. when two or more people are simultaneously attending to the same object), understand the social dynamics within the group and gather information about the event space itself.

Joint FOV localization can be performed by matching the videos taken from one POV device with the videos taken from another POV device. If there are $n$ people in the group, $P=\left\{p_i|i \in [1,n]\right\}$, then we have $n$ POV videos: $V=\left\{v_i|i \in [1,n]\right\}$. In the first step, all the videos in $V$ are synchronized by time-stamp. In the second step, $k$ videos (where $k \leq n$) are chosen from $V$ and matched against each other, which results in a total of $\binom{n}{k}$ matches. Matching is done frame-by-frame, by treating frame from one video as $I_{pov}$ and the frames from the other videos as $I_{ref}$. By thresholding the egocentric FOV localization scores, we can discover regions in time when the $k$ people were jointly paying attention to the same object. Finally, in the third step, the videos can be matched against the reference imagery from the event space to find out \emph{what} they were jointly paying attention to.

\begin{figure*}[t]
\begin{center}
\includegraphics[width=1.8\columnwidth]{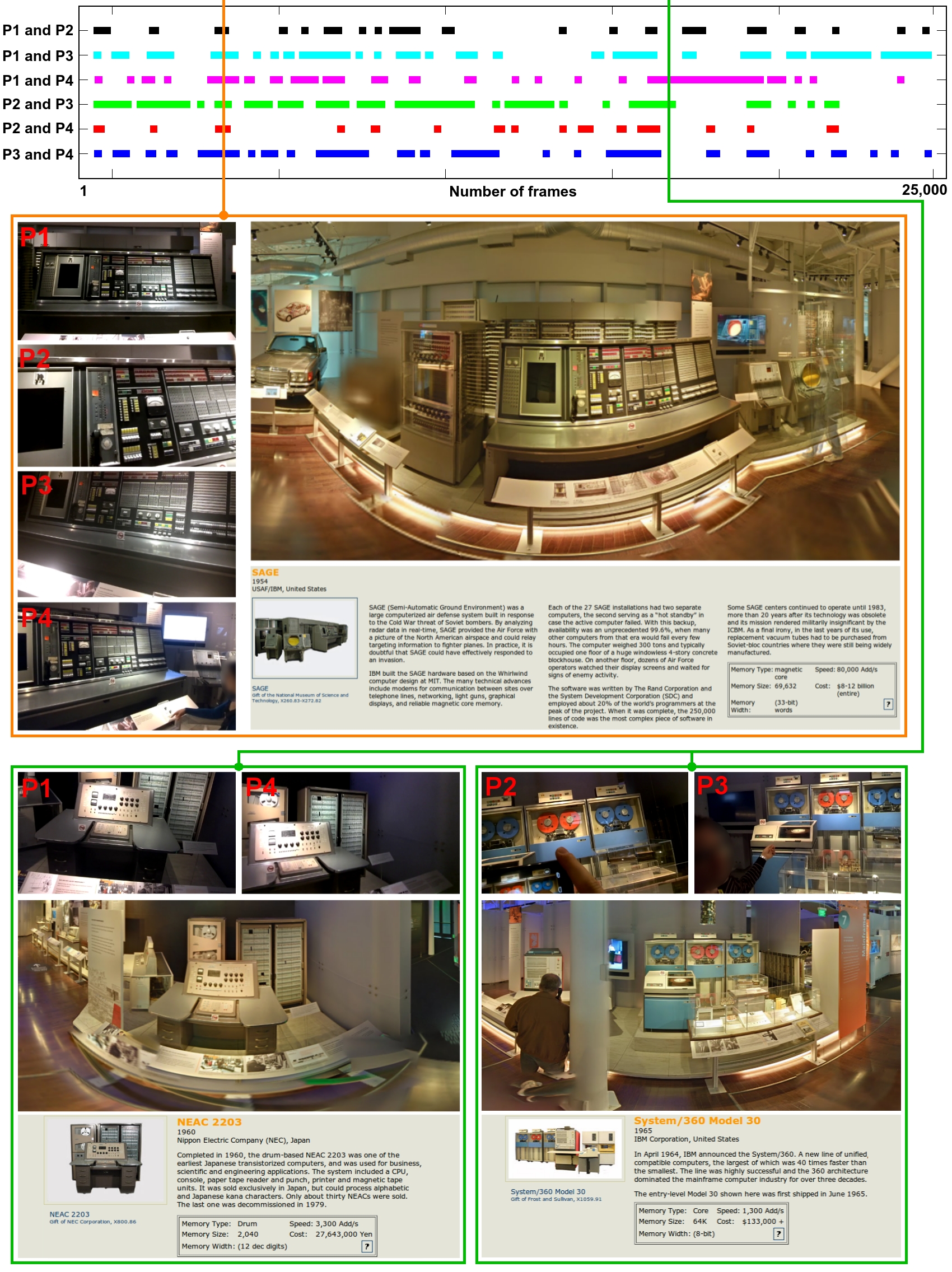}
\end{center}
\caption{\label{fig:joint_attention}The plot on the top shows the joint egocentric attention between groups of people. The x-axis shows the progression of time, from frame 1 to frame 25,000. Each row shows the result of joint egocentric FOV localization, i.e. the instances in time when pairs of people were jointly paying attention to the same exhibit in the museum. The orange vertical line indicates an instance in time when all the people (P1, P2, P3 and P4) were paying attention to the same exhibit. The green vertical line indicates an instance in time when P1 and P4 were jointly paying attention to an exhibit while P2 and P3 were jointly paying attention to a different exhibit. The corresponding frames from their Glass videos is shown. When matched to the reference street view images, we can discover the exhibits that the groups of people were viewing together and were probably having a discussion about. Details of the exhibit was automatically fetched from the reference dataset's annotation.}
\vspace{-0.75em}
\label{fig:chm}
\end{figure*}

We conducted our experiments with $n=4$ participants. The 4 participants wore Glass and visited the Computer History Museum in California. They were instructed to behave naturally, as they would on a group outing. They walked around in the museum looking at the exhibits and talking with each other. A total of 60 minutes of POV videos and the corresponding head-orientation information were captured from their 4 Glass devices. The videos are 720p at 30fps. The reference dataset consists of indoor street view panoramas from the museum. Next, joint egocentric FOV localization was performed by matching pairs of videos against each other, i.e. $k=2$, for a total of 6 pairs of matches. Figure \ref{fig:joint_attention} shows the results for 25,000 frames of video for all the 6 match pairs. The plot shows the instances in time when groups of people were paying attention to the same exhibit. Furthermore, we get an insight into the social dynamics of the group. For example, we can see that P2 and P3 were moving together but towards the end P3 left P2 and started moving around with P1. Also, there are instances in time when all the pairs of videos match which indicates that the group came together as a whole. One such instance is highlighted in Figure \ref{fig:joint_attention} by the orange vertical line. There are also instances when the 4 people split into two groups. This is shown by the green vertical line in Figure \ref{fig:joint_attention}.

\begin{figure}
\begin{centering}
\includegraphics[width=0.9\columnwidth]{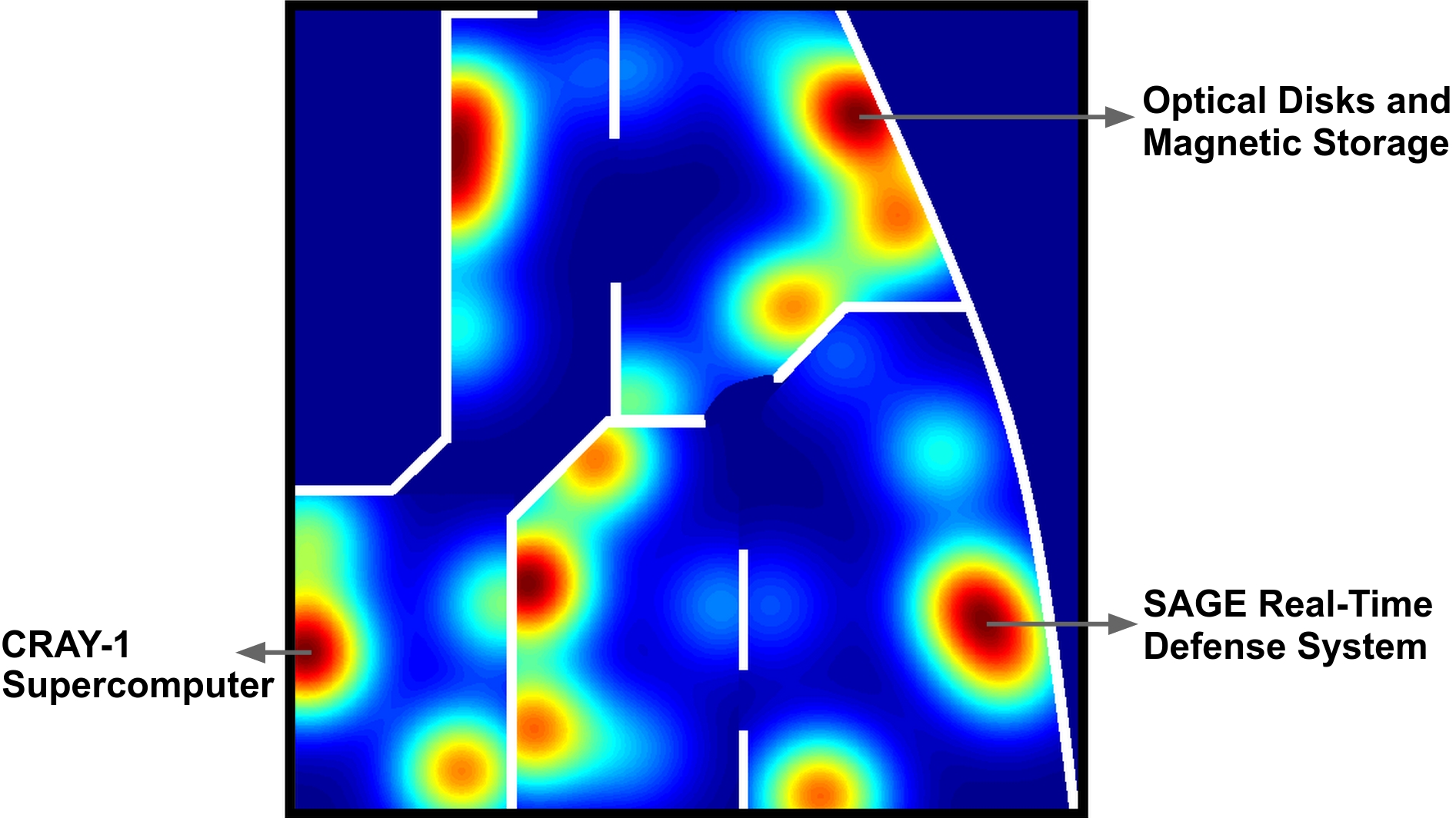} 
\par\end{centering}
\caption{\label{fig:heatmap}A heatmap overlaid on a section of the Computer History Museum's floorplan. Hotter regions in the map represent exhibits which had joint egocentric attention from more people. Three of the hottest regions are labeled to show the underlying exhibits that brought people together and probably led to further discussions among them.}
\vspace{-1.5em}
\end{figure}

Joint egocentric FOV localization also helps us get a deeper understanding of the event space. Interesting exhibits tend to bring people together for a discussion and result in higher joint egocentric attention. It is possible to infer this from the data by matching the videos with the reference images and labeling each exhibit with the number of people who jointly viewed it. By overlaying the exhibits on the floorplan, we can generate a heat map of the exhibits where hotter regions indicate more interesting exhibits that received higher joint attention. This is shown in Figure \ref{fig:heatmap}. Getting such an insight has practical applications in indoor space planning and the arrangement and display of exhibits in museums and other similar spaces.

\section{Discussion}

One of the assumptions in the paper is the availability of reference images in indoor and outdoor spaces. This may not be true for all situations. Also, it may not be possible to capture reference data concurrently (as in the indoor presentation dataset) due to restrictions by the event managers and/or privacy concerns. However, our assumption does holds true for a large number of indoor and outdoor spaces which makes the proposed approach practical and useful.

There are situations where the proposed approach may fail. While our matching pipeline is robust to a wide variation of changes in the images, it may still fail if the reference image is drastically different from the POV image (for example, a POV picture taken in summer matched against a reference image taken on a white snowy winter). Another reason for failure could be when the reference dataset is outdated. In such scenarios, the POV imagery will not match well with the reference imagery. However these drawbacks are only temporary. With the proliferation of cameras and the push to map and record indoor and outdoor spaces, reference data for our approach will only become more stable and reliable.

Our reference images are 2D models of the scene (for example, Street View panoramas). Moving to 3D reference models could provide a more comprehensive view of the event space and result in better FOV localization. But this would require a computationally intensive matching pipeline which involves 2D to 3D alignment and pose estimation.

\section{Conclusion}

We have demonstrated a working system that can effectively localize egocentric FOVs, determine the person's point-of-interest, map the shifts in FOV and determine joint attention in both indoor and outdoor environments from one or more POV devices. Several practical applications were presented on ``in-the-wild'' real-world datasets.

{\small
\bibliographystyle{ieee}
\bibliography{WACV2015}
}

\end{document}